\documentclass[journal]{IEEEtran}

\usepackage[T1]{fontenc}
\usepackage{amsmath}
\usepackage[cmintegrals]{newtxmath}
\usepackage{times}
\usepackage{verbatim}
\usepackage{color}
\usepackage{url}
\usepackage{graphicx}
\usepackage{array}
\usepackage{lmodern}
\usepackage{color}
\usepackage{colortbl}

\usepackage{amsmath}
\usepackage{subfig}
\usepackage{multirow}
\usepackage{amssymb}
\usepackage{wallpaper}
\usepackage{graphics,graphicx,pdfpages}
\usepackage{textcomp}
\usepackage{array}
\newcolumntype{P}[1]{>{\centering\arraybackslash}p{#1}}

\ifCLASSINFOpdf
\else
\fi

\begin{document}
%
\title{We Can ``See'' You via Wi-Fi --- WiFi Action Recognition via Vision-based Methods}
%
%
%

\author{Jen-Yin Chang$^{\star}$, Kuan-Ying Lee$^{\star}$, Yu-Lin Wei, Kate Ching-Ju Lin$^{\dagger}$, Winston Hsu\\
National Taiwan University, Taipei, Taiwan $^{\dagger}$National Chiao Tung University, Hsinchu, Taiwan\\
$^{\star}$Co-primary authors}

\maketitle

\begin{abstract}
Recently, Wi-Fi has caught tremendous attention for its ubiquity. Motivated by Wi-Fi's low cost and privacy preservation, researchers have been putting lots of investigation into its potential on action recognition and even person identification. In this paper, for bringing a new modality for multimedia community, we offer an comprehensive overview on these two topics in Wi-Fi.
Also, through looking at these two topics from an unprecedented perspective, we could achieve generality instead of designing specific ad-hoc features for each scenario. Observing the great resemblance of Channel State Information (CSI, a fine-grained information captured from the received Wi-Fi signal) to texture, we propose a brand-new framework based on computer vision methods.
To minimize the effect of location dependency embedded in CSI, we propose a novel de-noising method based on Singular Value Decomposition (SVD) to eliminate the background energy and effectively extract the channel information of signals reflected by human bodies. From the experiments conducted, we demonstrate the feasibility and efficacy of the proposed methods. Also, we conclude the factors that would affect the performance and highlight a few potential issues that require further deliberations.
\end{abstract}


%
\IEEEpeerreviewmaketitle

\section{Introduction}
%
%
%
%

\IEEEPARstart{I}{dentifying} who is performing which action has been deemed as one of the most important multimedia applications. For example, with the emergence of smart space, sensing human actions or activities automatically becomes essential. Moreover, with identification, user's information can be used to trigger specific customization such as temperature or illumination adjustment in room, content shown on television etc.

\begin{figure}[t]
\centering{
\resizebox{\columnwidth}{!}{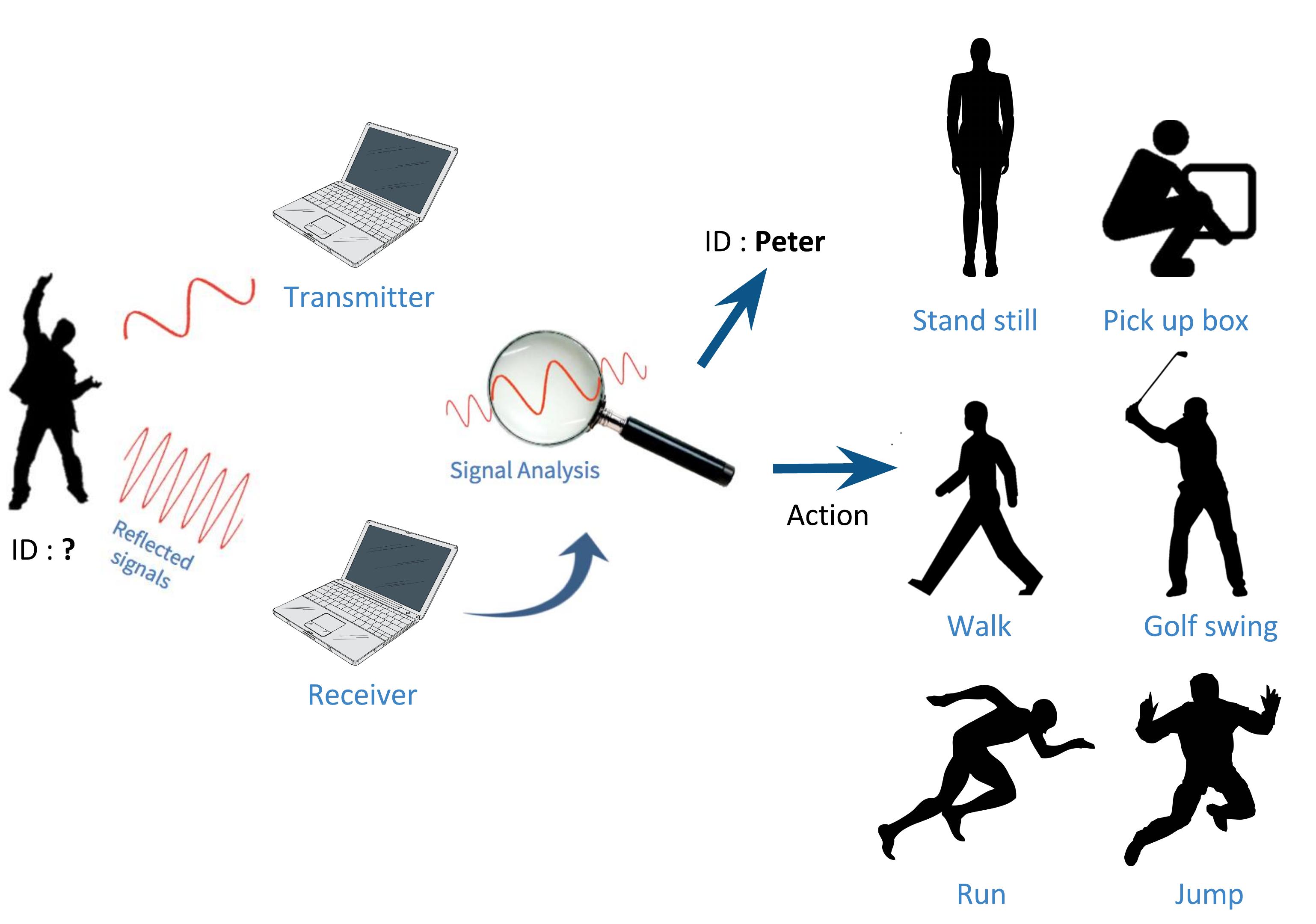}
\caption{Wi-Fi sensing scenario. We analyze the change of Wi-Fi signal caused by human action, classify it into a predefined action set, and identify the person performing it.}
\label{fig:scenario}
}
\end{figure}

\begin{figure*}[t]
\centering{
\resizebox{\textwidth}{!}{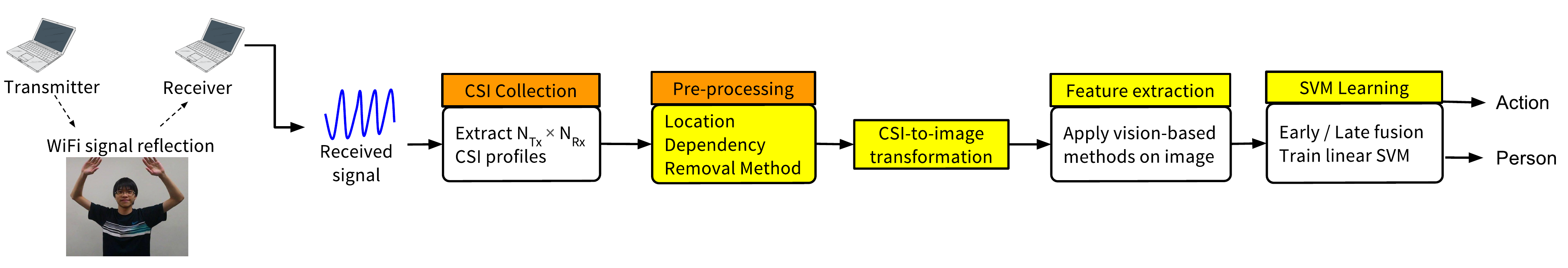}
\caption{Framework Overview. Different from image/video-based methods, we utilize Wi-Fi signals for recognition. (Blocks with yellow mark are the main differences between other works and our work.)}
\label{fig:system}
}
\end{figure*}

\par Several action recognition methods have been deployed, such as learning trajectory \cite{nn-traj}, motion descriptor \cite{racket-sport}, or optical flow and gradient descriptors \cite{codebook}. Overall, an 90\% accuracy has been achieved among most vision-based works. However, cameras might not be applicable in every place and scenario. For example, in restroom where privacy is of the first priority or in places where lighting is scarce, cameras are of little use. Nevertheless, action recognition could not be spared in these places. For instance, timely detection of falling in a bathroom limits the damage to minimum. Hence, previous works propose using wearable devices such as accelerometers to obtain the speed profile and detect the action \cite{accelerometer}. Those solutions, however, require users to wear devices all the time, which is unrealistic in some scenarios like taking a shower. Hence, one might prefer a non-contact method, avoiding intruding in users' daily life. To achieve identification, the most intuitive and popular way is face recognition. Though, a pretty high accuracy has been achieved with the power of deep learning \cite{deepface}, most works analyze on frontal faces, and in fact, side-view recognition is still very challenging. Some researchers propose to identify a person through gait analysis from sensor data. Yet, as the problem mentioned above, users would prefer a device-free solution. Given the flaws of the aforementioned methods, Wi-Fi leaps out at researchers' radar.

\par Preliminary research on RF-based action recognition collects raw signals received by USRP~\cite{usrp}, a software defined radio, and tracks the timestamps of motion \cite{wisee, allsee}. However, commercial APs do not provide raw signals, urging later works \cite{shopper, eeye} to seek for a more practical solution that analyzes on CSI obtained from the modified driver \cite{csitool}. We also realize our work based on this CSI toolkit. Though extensive efforts have been put into CSI-based action recognition, to the best of our knowledge, most existing works extract ad-hoc features which might encounter accuracy degradation as environments change. On the other hand, prior work on identification \cite{rfcapture} utilizes antenna array along with tailored devices to recognize the person standing behind the wall, triggering us to achieve identification via commercial APs. 

\par We, interestingly, observe that a CSI-transformed image actually has some special texture appearances. Hence, in this work, we also use the CSI toolkit, but aim at investigating a general vision-based solution for action recognition and user identification. The fundamental challenge of such a vision-based method is how to exclude the effect of location dependent information typically embedded in the captured CSI. To overcome this issue, we propose a de-noising method based on SVD to improve reliability of cross-environment performance. We further extend our vision-based framework to identify users, and verify its accuracy for various distances, packets sampling rates and outfits. Our contributions include:
\begin{itemize}
  \item We provide an overview on action recognition and person identification via Wi-Fi.
  \item To our best knowledge, we are the first to investigate the feasibility of processing CSI by vision-based methods with extendible learning-based framework \cite{icassp}.
  \item Despite of the promising performance in one room, we enable cross-environment action recognition by removing location dependency via the SVD-based de-noise method \cite{mm}.
  \item We further experimentally verify the possibility of using our vision-based framework to identify users, and discuss some factors that may affect performance and are worth future studies as open problems.
\end{itemize}

\begin{table*}[h!]
\begin{center}
\resizebox{\textwidth}{!}{
\begin{tabular}{|c|P{65pt}|P{48pt}|P{78pt}|P{180pt}|}
\hline
  Work & Software-Defined Radio & Directional Antenna & Fequency Modulated Carrier Wave & Task\\ \hline
  WiSee \cite{wisee}, AllSee \cite{allsee} & V & V &  & Coarse-grained action recognition \\
  E-eyes \cite{eeye}, CARM \cite{carm} &  &  &  & Fine-grained action recognition \\
  RF-Capture \cite{rfcapture} &  & V & V & Identification \\
  WiTrack \cite{witrack} &  & V & V & Human body localization \\ \hline
  Our solution &  &  &  & Fine-grained action recognition \& Identification \\
\hline
\end{tabular}
}
\caption{Comparison of specified hardware among related works. 
The mark ``V'' means the corresponding work utilizes the instruments or techniques listed above in the header, while ``Task'' in the header indicates what the work aims to solve.
}
\label{tab:compare}
\end{center}
\end{table*}


\section{Related Works}
\par In this section, we introduce some related works using wireless signals with divergent aspects. 

\subsection{Raw Signal}
\label{ssec:raw}
\par From the frequency aspect, if we view human body as a source of reflected signals, when the user pushes toward a receiver, the relatively approaching speed causes a positive Doppler shift at the receiver. On the contrary, a negative Doppler shift occurs as the user's hand departs from the receiver. Harnessing the Doppler effect, WiSee~\cite{wisee} achieves a 94\% accuracy differentiating between nine gestures. As from the amplitude angle, since the total path from a transmitter to human body and to a receiver is shorten as pushing happens, the power dissipation decreases, rendering a rising amplitude on the receiver side. Utilizing this phenomenon, AllSee~\cite{allsee} successfully reduces the computational cost by performing analysis directly on the time domain signals with an 91\% accuracy classifying four gestures. However, such raw signals are only available on special hardware such as USRP
or WARP~\cite{warp}, nudging researchers toward using CSI, which can, in practice, be accessed from today's commodity devices and computed by the modified driver \cite{csitool}.

\subsection{Channel State Information}
\label{ssec:csi}
\par With CSI, more actions and even human activities are successfully recognized, giving rise to many interesting applications.  E-eyes~\cite{eeye} presents a user-feedback system, separating actions into walking and in-place activity, which is capable of identifying several trajectories and activities. WiHear~\cite{wihear} processes CSI for mouth motion profiles in order to read what people say. CARM~\cite{carm} leverages a de-noising method based on principal component analysis (PCA) followed by discrete wavelet transform (DWT), and supports human activity recognition independent of environment variances. Nonetheless, parts of the features used in CARM~\cite{carm} are related to the time duration of an action, which to our knowledge, might render classifiers highly vulnerable to duration estimation errors. Also, despite the promising results, most works involve ad-hoc domain knowledge specifically related to the defined scenarios such as WiFall~\cite{wifall} primarily designed for fall detection.

\par And with its high sensitivity to environmental variances caused by moving objects, researchers start exploring the possibility of Wi-Fi identification. In FreeSense~\cite{freesense}, Tong et al. propose an approach that identifies the line-of-sight path crossing moments followed by PCA and DWT to extract features for K-nearest neighbor classifier and achieve an accuracy of 90\% in a six-person scenario. Yet, their approach would work only if a subject passes through the LOS path. WifiU~\cite{WifiU} also harnesses PCA to reduce the uncorrelated noises in different subcarriers. By applying short-time Fourier transform to convert PCA components into spectrograms, it achieves a top-1 accuracy of 92\%. However, several of their features such as walking speed, gait cycle time, to the best of our knowledge, would be ineffective when applied on actions other than walking. For instance, aperiodic and in-situ actions like jump do not involve movement speed or period.

\par Also, different to these works, we focus on providing a general feature that could not only be applied on identification even when a person performs actions other than walking, but also be utilized in different objectives, such as action recognition.

\subsection{Others}
\par Besides CSI from commercial APs, some researchers utilize directional antennas or antenna array to extract more detailed information, achieving a better localization resolution and realizing more powerful applications. RF-Capture~\cite{rfcapture} designs a unique device using a T-shape antenna array and Frequency Modulated Carrier Wave (FMCW) to identify persons through wall. The antenna array differentiates the direction of signals and the frequency chirp estimates the delay of received signals by constantly tuning the transmitted signals. WiTrack~\cite{witrack} uses a similar device to localize a person in 3D space through wall. These works are orthogonal to our work. Requiring simply commercial APs with only omni-directional antennas, which are rather low-cost and ubiquitous, our framework can be deployed widely. 

\subsection{Comparison}
\par We conduct a comprehensive survey into recent works harnessing FMCW radar, directional antenna, software-defined radio and commercial AP in Table \ref{tab:compare}, comparing the main differences among these works in order to differentiate our works with other researches. 

\par
\begin{figure}[t]
\centering{
\resizebox{\columnwidth}{!}{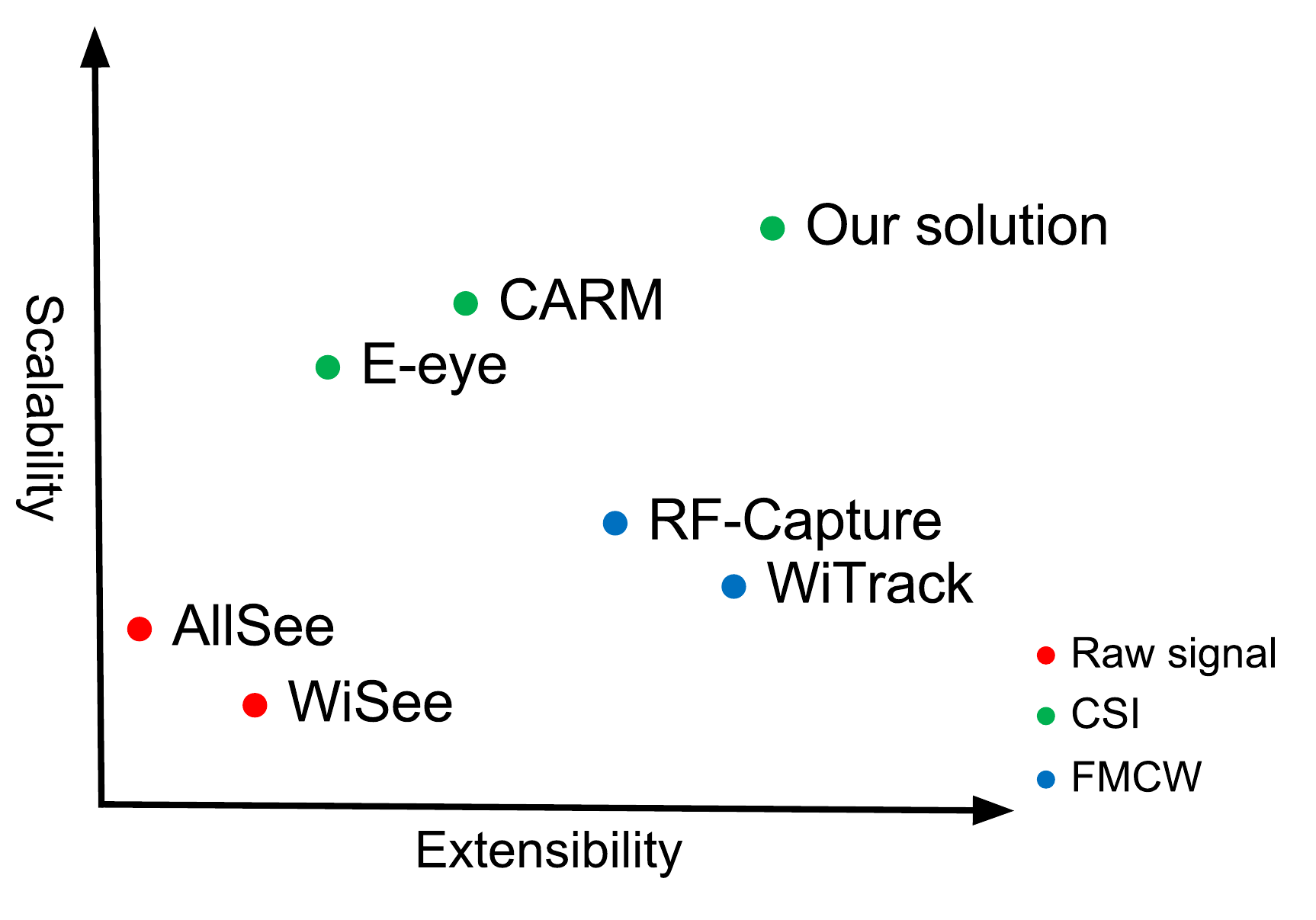}
\caption{Design space comparing to related works. Scalability means the easiness to deploy the proposed framework, considering hardware specificity and accessibility. Extensibility is how well a framework could be extended and since our proposed framework takes advantage of vision techniques, it could be rather easy to take in any promising vision works, making it flexible.}
\label{fig:scenario_diff}
}
\end{figure}

 We further place related works on a quadrantal diagram with the two axes being scalability and extensibility in Figure \ref{fig:scenario_diff}. Scalability measures how well a framework could be spread while extensibility measures how well a framework could be adapted to new techniques. The reason we choose these two aspects among numerous designing criteria is we aim at designing a system that could be easily deployable. 

\par For scalability, since E-eye~\cite{eeye}, CARM~\cite{carm} and our method resort to commercial off-the-shelf instruments such as AP and laptops, we consider them easier to scale up. Since WiSee~\cite{wisee}, AllSee~\cite{allsee}, RF-Capture~\cite{rfcapture} and WiTrack~\cite{witrack} utilize specifically designed hardware or software defined radios, we consider rather difficult and hence place them in the lower position of the figure. For extensibility, since WiTrack~\cite{witrack}, RF-Capture~\cite{rfcapture} and ours take advantage of learning algorithms rather than designing dedicated ones such as WiSee~\cite{wisee} or AllSee~\cite{allsee}, we consider them to be more extensible as novel learning techniques come out. Hence, we put them on the right side of the graph. 
 

\section{Channel State Information}
\par In this section, we deeply dig into what CSI is, how it could be used for vision-based methods and describe the location dependency embedded in CSI.

\subsection{Background}
\label{sec:background}
\par As Wi-Fi waves propagate from a transmitter (Tx) through the air to a receiver (Rx), it bumps into objects and goes through several reflection, absorption, scattering or other effects caused by human or objects before reaching the receiver (Rx). When an action takes place, reflecting paths associated with human body differ. Action recognition could thus be realized by observing the variation of the CSI. Current Wi-Fi transmitting protocols such as 802.11n implement OFDM (Orthogonal Frequency-Division Multiplexing) for reducing interference and fading. It segments the bandwidth into several closely-spaced sub-carriers, each carrying a data stream. For more details about OFDM please refer to \cite{ofdm}. As receiving signals, a receiver learns its Channel State Information (CSI), which is a complex number detailing the phase shift and power decay corresponding to the decay and propagation delay of multiple paths.
In particular, we can formulate the communication model as $Y = HX$, where $X$ is the transmitted signals, $Y$ is the received signals, and $H$ is the CSI, denoting the overall influence of the environment to $X$. After decoding the received preambles or pilot bits, which are information both known by Tx and Rx, the receiver can learn the CSI, i.e., $H$, by comparing the differences between $Y$ and $X$. When an action happens, the number of reflecting paths and their distances change accordingly and thereby, by extracting information from the CSI, one could classify the action performed.

\subsection{Formulation}
\label{ssec:cfr}
From CARM~\cite{carm}, we know the amplitude of a sub-carrier with frequency \textit{f} at time \textit{t} could be expressed as \textit{H(t, f)} in equation (\ref{eq:cfr-simple}), where \textit{H\textsubscript{s}(f)} is the aggregate channel of all the static paths, \textit{H\textsubscript{d}(f, t)} is the aggregate channel of the dynamic paths and \(\Delta f\) is the frequency offset between Tx and Rx.

\begin{equation} \label{eq:cfr-simple}
H(f, t) = e^{-j2\pi\Delta ft}\left(H_s(f) + H\textsubscript{d}(f,t)\right)
\end{equation}
\(H\textsubscript{d}(f) = \sum_{k\in P_d}^{}a_k(f, t)e^{-j\frac{2\pi d_k(t)}{\lambda}}\), where \textit{a\textsubscript{k}(f, t)} is attenuation of the \textit{k\textsubscript{th}} path at time \textit{t} and frequency \textit{f}, \textit{d\textsubscript{k}(t)} is the distance of \textit{k\textsubscript{th}} path and \textit{P\textsubscript{d}} is the set of all dynamic paths. 
 
\par If an object moves at a constant speed, the distance of the \textit{k\textsubscript{th}} path, \textit{d\textsubscript{k}(t)}, could be expressed as \textit{d\textsubscript{k}(t) = d\textsubscript{k}(0) + v\textsubscript{k}}. Thus the power of CSI \(|H(f, t)|^2\) at time \textit{t} and frequency \textit{f} can be derived as:

\begingroup\makeatletter\def\f@size{7.5}\check@mathfonts
\def\maketag@@@#1{\hbox{\m@th\large\normalfont#1}}
\begin{align} \label{eq:cfrcomplex}
&|H(f, t)|^2= \\
&\sum_{k\in P_d}2|H_s(f)a_k(f, t)|\cos\left(\frac{2\pi v_kt}{\lambda} + \frac{2\pi d_k(0)}{\lambda} + \phi_{sk}\right) + C(f,t) + \nonumber \\
&\sum_{\substack{k,l\in P_d \\ k\neq l}}2|a_k(f, t)a_l(f, t)|\cos\left(\frac{2\pi (v_k - v_l)t}{\lambda} + \frac{2\pi (d_k(0) - d_l(0))}{\lambda} + \phi_{kl}\right), \nonumber
\end{align}
\endgroup
where \(C(f, t)\) is a constant given sub-carrier frequency $f$ and time $t$, and \(\phi_{sk}\) and \(\phi_{kl}\), respectively, represent initial phase offsets of transmitter and receiver.

\par
\begin{figure}[h]
\centering{
\resizebox{\columnwidth}{!}{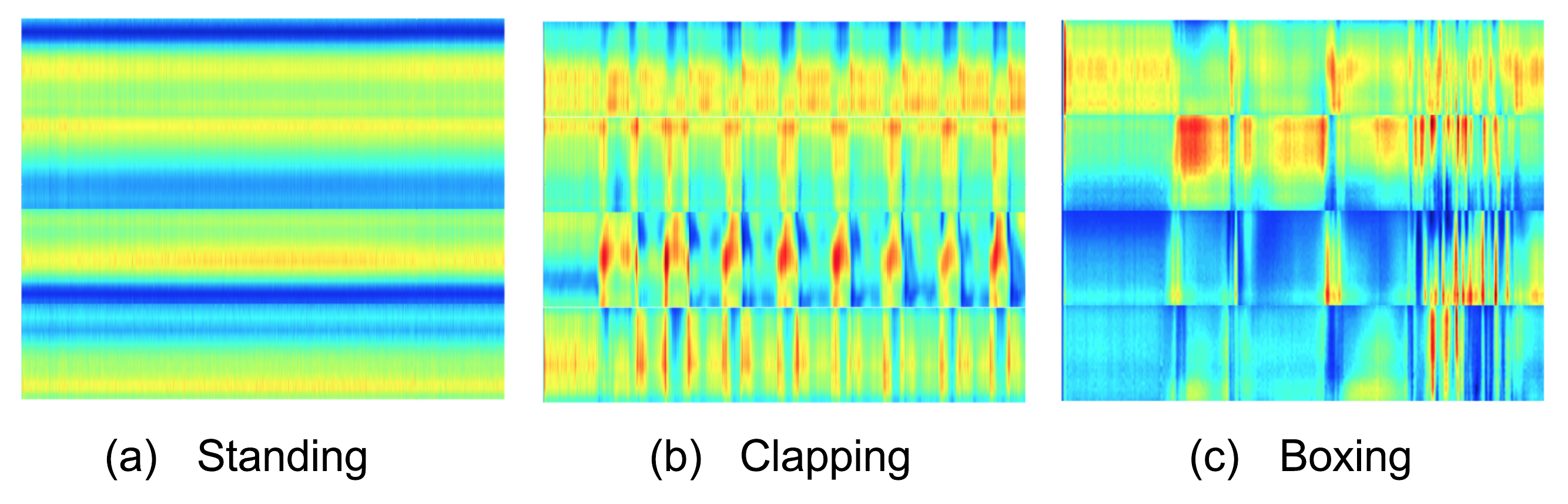}
\caption{Transformed images of (a) standing, (b) clapping, (c) boxing (X-axis: timestamp, Y-axis: 30 sub-carriers \(\times\) 4 channels). We could observe a faster punching speed leads to denser stripes, as in the rear part of boxing.}
\label{fig:cfr}
}
\end{figure}

 We observe that in equation (\ref{eq:cfrcomplex}), frequencies of cosine waves are determined by the action speed \textit{v\textsubscript{k}}. A faster speed leads to a larger phase change and renders denser stripes on transformed images, as shown in Figure \ref{fig:cfr}. Since actions of different speeds present different textures on transformed images, we propose applying vision-based methods on transformed images.

\begin{figure}[t]
\centering{
\resizebox{\columnwidth}{!}{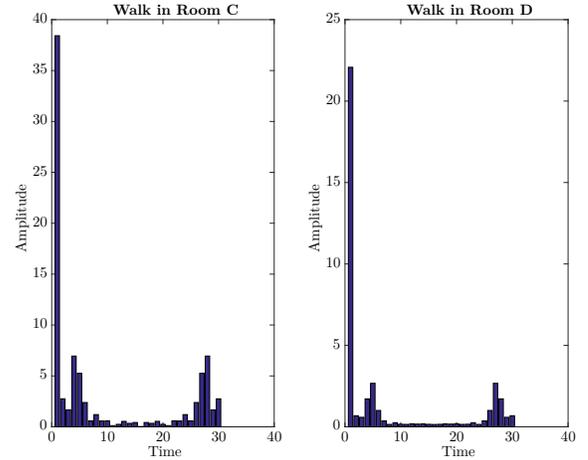}
\caption{Each tap represents the aggregated time-domain channel of the
paths with similar distances. A smaller time-domain index corresponds
the signals from shorter paths, thereby causing a shorter delay.} 
\label{fig:ifft_result}
}
\end{figure}

\subsection{Location Dependency}
\par To elaborate on why CSI embeds location information, we transform a
frequency-domain CSI profile collected from our testbed to the time
domain by inverse fast Fourier transformation. As shown in Figure
\ref{fig:ifft_result}, the first tap usually represents the channel
of the line-of-sight path (with the shortest delay), and, hence, has
the strongest amplitude. The rest of weaker taps consist of other
static paths, reflected by nearby objects, and dynamic paths,
reflected by human bodies. We can observe that, even after removing
the strongest line-of-sight path, the remaining signal patterns in the
two rooms are still very different because the reflected static paths
change with environments and, more importantly, still have a
relatively strong power, as compared to those reflected by human.

\par Hence, we argue that, to enable location-independent recognition, we
should remove those dominant environmental information and only
transform the CSI of dynamic paths into images.

\section{Location Dependency Removal Method}
\par To deal with location dependency, we propose a method based on SVD to remove the background energy.

\par In CARM~\cite{carm}, Wei et al. consider the correlation between CSI
streams and point out the fact that any effects to signals would be
presented across CSI streams, inspiring us to apply SVD on CSI. In
image processing techniques, SVD is an attractive algebraic transformation
used to manipulate an image in two distinctive sub-spaces, i.e., data
spaces and noise spaces. \cite{svd} If we treat the CSI profile as a matrix $H$ of dimension $t \times N\textsubscript{subcarriers}$ (denoted as $d$), after applying SVD, $H$ is factorized into three matrices, $U, S$ and $V$, where $U$ is a $t \times t$ orthogonal
matrix, $V$ is a $d \times d$ orthogonal matrix, and $S$ is a $t\times d$ matrix with the diagonal elements representing the singular
values of $H$.

\par We can formulate low rank approximation of $H$ as follows:
%
\begin{align} \label{eq:compress}
&H' = \sum_{i =
1}^{k}s_i
u_i
v_i^T =
s_1u_1v_{1}^T +
s_2u_2v_{2}^T +
\cdots + 
s_ku_kv_{k}^T,
\end{align}
%
in which $k$ is usually smaller than $d$ for compression and $s_i$
means the singular values (SV) in decreasing order corresponding to a
pair of singular vectors $u_i$ and $v_i$. Intuitively, this equation
shows that $H$ is a sum of different bases with its importance score
for approximating the original $H$.

\par
\begin{figure}[t]
\centering{
\resizebox{\columnwidth}{!}{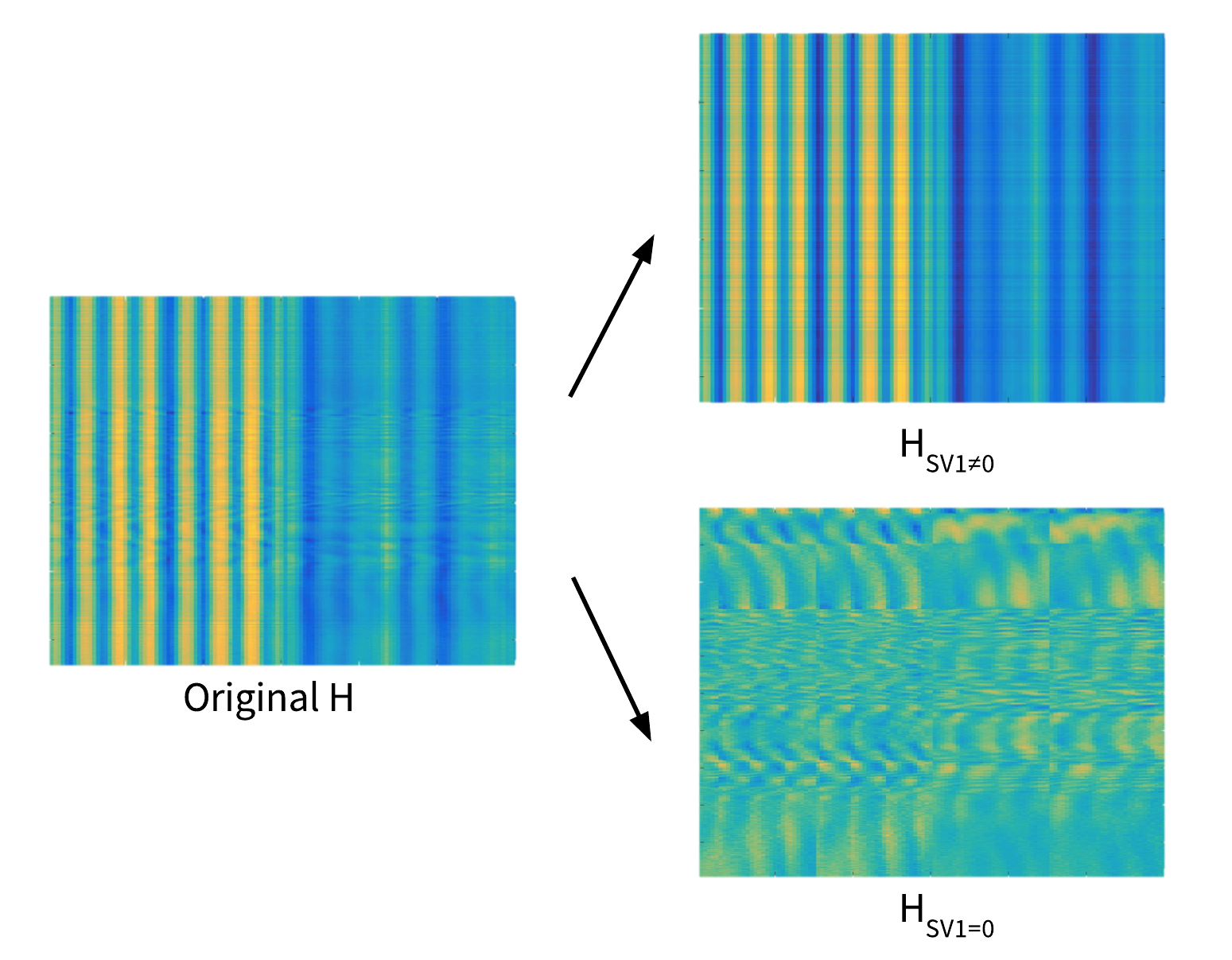}
\caption{CSI-transformed images of picking up box from different
reconstructed channels $H$.}
\label{fig:svd_example}
}
\end{figure}

 After decomposition, we remove the dominant background energy by setting the first singular value to zero for each image. We illustrate our design rationale via the example shown in Figure \ref{fig:svd_example}. We can observe from the figure that if
we set all SV in $S$ to zero except for the largest one, denoted by
$S=S_{SV_1 \neq 0}$, and reconstruct the channel information by
$H_{SV_1\neq 0}=US_{SV_1 \neq 0}V^T$, then, the reconstructed channel
$H_{SV_1\neq 0}$ would be similar to the CSI recorded in the quite
environment. This example shows that the largest SV packs
most of energy corresponding to the static paths. Therefore, we set
the largest SV to zero so that the background can be mostly removed
and the reconstructed channel $H_{SV_{1}=0}$ preserves only the human
action part.

\par
\begin{figure}[t]
\centering{
\resizebox{\columnwidth}{!}{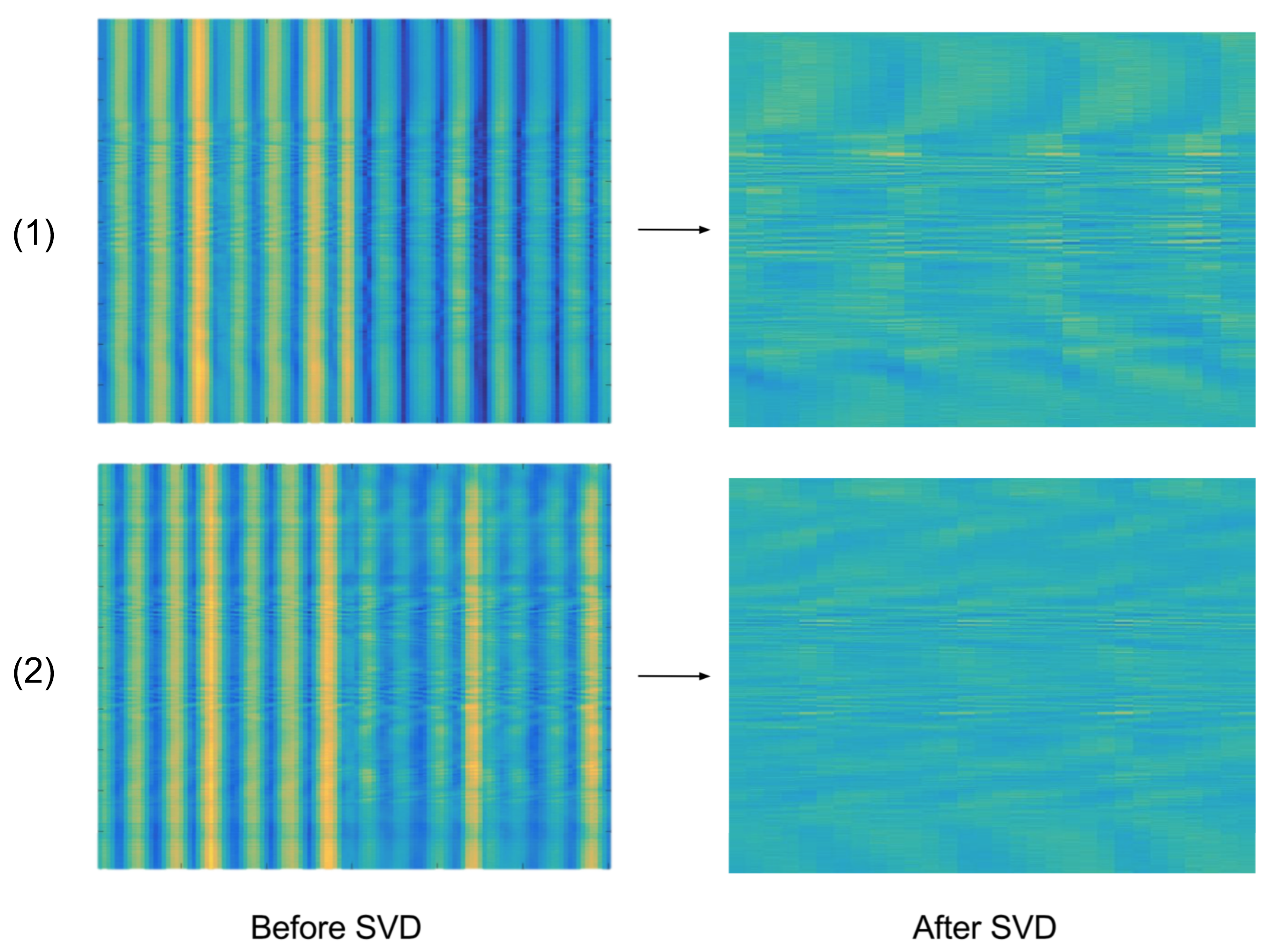}
\caption{CSI-transformed images of the same action (pick up box) from two different rooms.}
\label{fig:beforeafter}
}
\end{figure}

 We illustrate how the SVD-based background removal method improves
the performance by the examples shown in Figure \ref{fig:beforeafter}. In this experiment, the subject performs the same action (pick up box) in two different rooms.
Before SVD is applied, even when the same action is performed, the
power distributions of the whole CSI captured in two locations differ due to distinct static
paths in different rooms, making the models trained in one place fail to recognize actions recorded in other places. However, after removing the largest SV, as shown in the right ones in Figure \ref{fig:beforeafter}, the textures of the remaining part from two different rooms resemble each other. 

\section{Proposed Vision-based Framework}
\par In this section, we describe the overall flow of the proposed framework, from collecting CSI, pre-processing, extracting features to training the classifier, as shown in Figure \ref{fig:system}.

\subsection{Collecting CSI}
\label{ssec:collectCSI}

\par We use MacBook Pro 2014 or ASUS U80v with Intel 5300 NIC as Tx and Fujitsu SH560 with Intel 5300 NIC as Rx, each having two antennas. With \textit{N\textsubscript{Tx}} = 2, \textit{N\textsubscript{Rx}} = 2, we have 2 \(\times\) 2 = 4 \textit{Tx-Rx} pairs, each generating a set of  CSI with dimension 30 (sub-carriers) \(\times\) \textit{t} (samples). We then process these four sets of CSI separately and investigate whether early or late fusion yields better performance. 


\subsection{Pre-processing}
\label{ssec:pre}
\par Due to interference caused by other devices in the same Wi-Fi channel, packets received are not evenly distributed in time. Thus, we linearly interpolate raw CSI to 1000 samples/second. We then apply 5th-order Butterworth filter with cutoff frequency 50Hz to remove high-frequency noise. And since power distributions of different sub-carriers vary, we normalize each sub-carrier by subtracting the average of a moving-window, width set as 300ms, from each sample.

\subsection{Location Dependency Removal}
\par We will later investigate, via experiments, the two settings: (1) apply SVD on each set of CSI with dimension $30 \times t$ separately, and (2) apply SVD on all the CSI streams with dimension $120 \times t$, and explore which option performs better.

\subsection{Feature Extraction}
\label{ssec:feature}
\par After transforming a set of CSI into an image of specific size, we experiment with Gabor and BoW-SIFT on it. Though deep features are potentially more powerful, due to the scarcity of current data we will not address it in our work.

\subsubsection{Gabor Filter}
\label{sssec:gabor}
\par A Gabor filter is defined by a plane wave multiplied by a Gaussian function. By setting different scales and orientations, a set of filters are obtained. (Details could be found in \cite{gabor}.)
These filters are convoluted with a transformed image. When a local patch resembles the filter, a high response will be obtained. Finally, a response map is produced, of which we then take two statistics, mean and standard deviation.

\par We set \textit{\#scale}, \textit{\#orientation} and size of the Gabor filters to 8, 6 and 15 respectively, which usually produce better accuracy from our measurements. Hence, the dimension of our final Gabor feature is 8 \(\times\) 6 \(\times\) 2 = 96. 

\subsubsection{Bag of Word-SIFT}
\label{sssec:sift}
\par SIFT (Scale Invariant Feature Transform) seeks to transform an image into a collection of keypoints, each described by a feature vector invariant to illumination, translation, rotation and scaling. \cite{sift}
We take all feature vectors of the training images from a \textit{Tx-Rx} pair and perform K-means clustering to find 48 centroids. BoW-SIFT feature is then generated by quantizing vectors of an image to the nearest centroid, producing a histogram of dimension 48.

\par Thus, in testing phase, we quantize the feature vectors of the input image into centroids found during training and feed the produced 48-dimension feature into the trained classifier.

\subsection{Training SVM Classifier}
\label{ssec:svm}
For each of the four \textit{Tx-Rx} pairs, we obtain a feature vector. We investigate fusing them before or after training linear SVM classifiers.

\subsubsection{Early Fusion} 
\label{sssec:earlyfuse}
We concatenate four features from four \textit{Tx-Rx} pairs into a new feature. Then, we train a single classifier and take action with the highest probability as the predicted result.

\subsubsection{Late Fusion} 
\label{sssec:latefuse}
Instead of concatenating four feature vectors and training a single classifier, we train one for each pair, so there would be four classifiers. Given a testing instance, probability of each action is obtained, rendering four probability vectors of length \#\textit{action} (seven in our case). Summing these four vectors, we take action with the highest probability as the predicted result.

\section{Experiments}\label{sec:exp}
In this section, we present our experimental results on action recognition and person identification.

\subsection{Action Recognition without SVD}
\subsubsection{Dataset and Settings}
\par
\begin{figure}[h]
\centering{
\resizebox{0.7\columnwidth}{!}{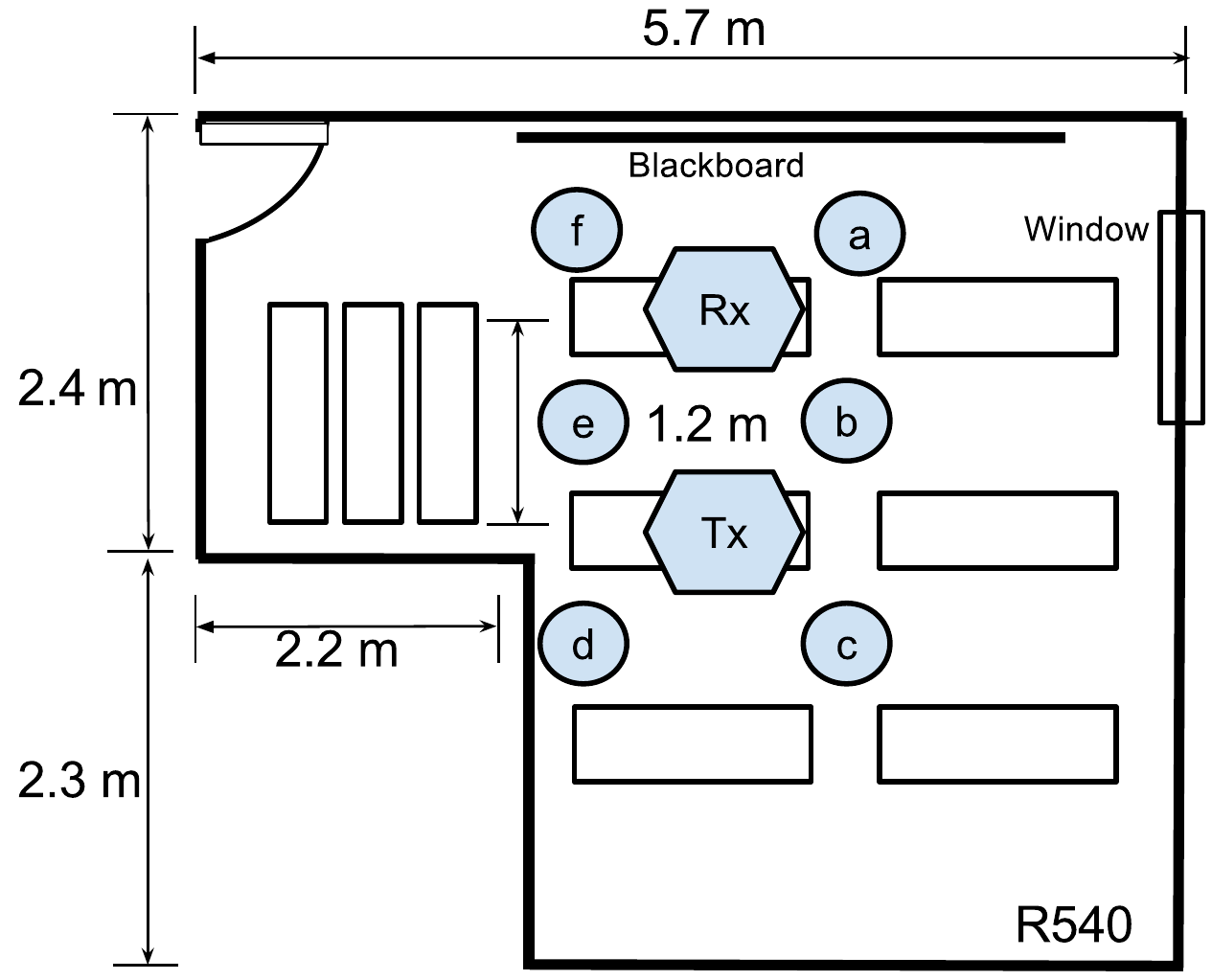}
\caption{Subject performs actions in each circle.}
\label{fig:seminar_room}
}
\end{figure}

 Dataset A is collected in a seminar room, as shown in Figure \ref{fig:seminar_room}, for verifying if our method could recognize actions as well as locations. We define seven actions: \textit{Box, Clap, Wave hand, Kick, Quick squat, Jump, Stand still} and six locations \textit{a, b, c, d, e, f}. A single subject performs each action 10 times on each location, so in total we have 7 \(\times\) 6 \(\times\) 10 = 420 data. Dataset B is collected to compare with vision-based methods on video action recognition, and thus we define actions the same as the benchmark dataset, KTH~\cite{KTH}. These actions include: \textit{Box, Clap, Wave hand, Walk, Jog, Run, Stand still}. Two subjects are asked to perform each action 10 times and in total we have 2 \(\times\) 7 \(\times\) 10 = 140 data.

\par All actions are performed in a 5-second period, each generating four sets of CSI with dimension 30 \(\times\) 5000 (interpolated to 1000 samples/second). We then transform them into four images of size 576 \(\times\) 432.

\subsubsection{Results}
\par We evaluate the performance using 10-fold cross validation. First, we conduct experiments on dataset A with late-fusion scheme (Details will be discussed in Table \ref{tab:late-early}), as cross-validation accuracy shown in Table \ref{tab:datasetA}. We could verify that viewing CSI as texture is feasible and Gabor filters, particularly suitable for texture recognition, perform better. Hence, following experiments are primarily based on Gabor.

\begin{table}[h!]
\begin{center}
\begin{tabular}{|c|c|c|}
\hline
  Location & BoW-SIFT & Gabor \\ \hline
  a & 64.29\% & 84.36\% \\ 
  b & 81.43\% & 96.57\% \\ 
  c & 77.14\% & 89.79\% \\ 
  d & 80.00\% & 86.71\% \\ 
  e & 51.43\% & 81.21\% \\ 
  f & 74.29\% & 81.50\% \\ \hline
  all & 50.48\% & 77.45\% \\ \hline
\end{tabular}
\caption{Accuracy of Gabor and BoF-SIFT on dataset A.}
\label{tab:datasetA}
\end{center}
\end{table}

\par Table \ref{tab:late-early} shows the results of cross-validation on both datasets using early fusion and late fusion. As the statistics show, late fusion performs better since it exploits four different channels with four classifiers. Though each classifier is weaker compared to that of early fusion, more channels provide more information for recognition.  Due to the superiority of late fusion to early fusion, we only list results of late fusion in the following experiments. Also from results on dataset B, we believe Wi-Fi could actually supports cameras in differentiating actions that are visually similar but of different CSI patterns.

\begin{table}[ht]
\centering
\begin{tabular}{|c|c|c|}
\hline
      & Dataset A   & Dataset B \\ \hline
  Early & 70.24\%   & 80.23\% \\
  Late  & 77.45\%   & 86.96\% \\ \hline
%
\end{tabular}
\caption{Accuracy of early and late fusion applying Gabor.}
\label{tab:late-early}
\end{table}

\par Wondering if location affects the accuracy, we split the classification process into two stages, namely, location identification followed by action recognition. Location is predicted using classifier trained on all data first.

\begin{table}[h!]
\begin{center}
\begin{tabular}{|c|c|}
\hline
  Target & Accuracy \\ \hline
  Location & 98.00\% \\
  Action (Given location) & 83.00\% \\
\hline
\end{tabular}
\caption{Accuracy of location-awareness classification.}
\label{tab:loc-aware}
\end{center}
\end{table}

\begin{table}[h!]
\begin{center}
\begin{tabular}{|c|c|c|}
\hline
  Training location & Testing location & Accuracy \\ \hline
   b & e & 12.86\% \\
   b, e & c & 17.14\% \\
   b, c, e & d & 42.86\% \\
   b, c, d, e & a & 23.00\% \\
   a, b, c, d, e & f & 17.00\% \\
\hline
\end{tabular}
\caption{Accuracy of cross-location classification.}
\label{tab:cross-loc}
\end{center}
\end{table}

\par Then a classifier trained with data of the suggested location is employed to classify the action, with results shown in Table \ref{tab:loc-aware}. The 98\% accuracy in predicting location and the 83.00\% accuracy in action classification (compared to 77.45\% in Table \ref{tab:late-early}) reveals that our features still embed location information. Hence, a classifier trained on all data performs poorly compared to a classifier dedicated for that specific location. To further testify, we experiment a trained classifier on data of unseen location with results demonstrated in Table \ref{tab:cross-loc}. We note though a slight performance boost might be witnessed as we incorporate data from more locations. When testing on an instance of a difference location, a classifier performs poorly and without consistency.


\par Finally, we conduct an experiment exploring whether size of the transformed images affects accuracy, as shown in Table \ref{tab:size}. The result demonstrates that as the size of images becomes smaller, performance remains excellent as long as the size of filters alters accordingly, implying the proposed framework is computationally efficient.

\begin{table}[h!]
\begin{center}
\begin{tabular}{|c|c|c|}
\hline
  Size        & Filter size   & Accuracy \\ \hline
  5000 \(\times\) 30  & 15    & 66.54\% \\
  576 \(\times\) 432  & 15    & \textbf{86.96\%} \\
  72 \(\times\) 54  & 15      & 75.64\% \\
  72 \(\times\) 54  & 9       & 84.46\% \\
\hline
\end{tabular}
\caption{Accuracy on dataset B between different sizes (in pixels) of image.}
\vspace{-10pt}
\label{tab:size}
\end{center}
\end{table}

\subsection{Action Recognition with SVD}
\subsubsection{Dataset and Settings}
\par
\begin{figure}[t]
\centering{
\resizebox{\columnwidth}{!}{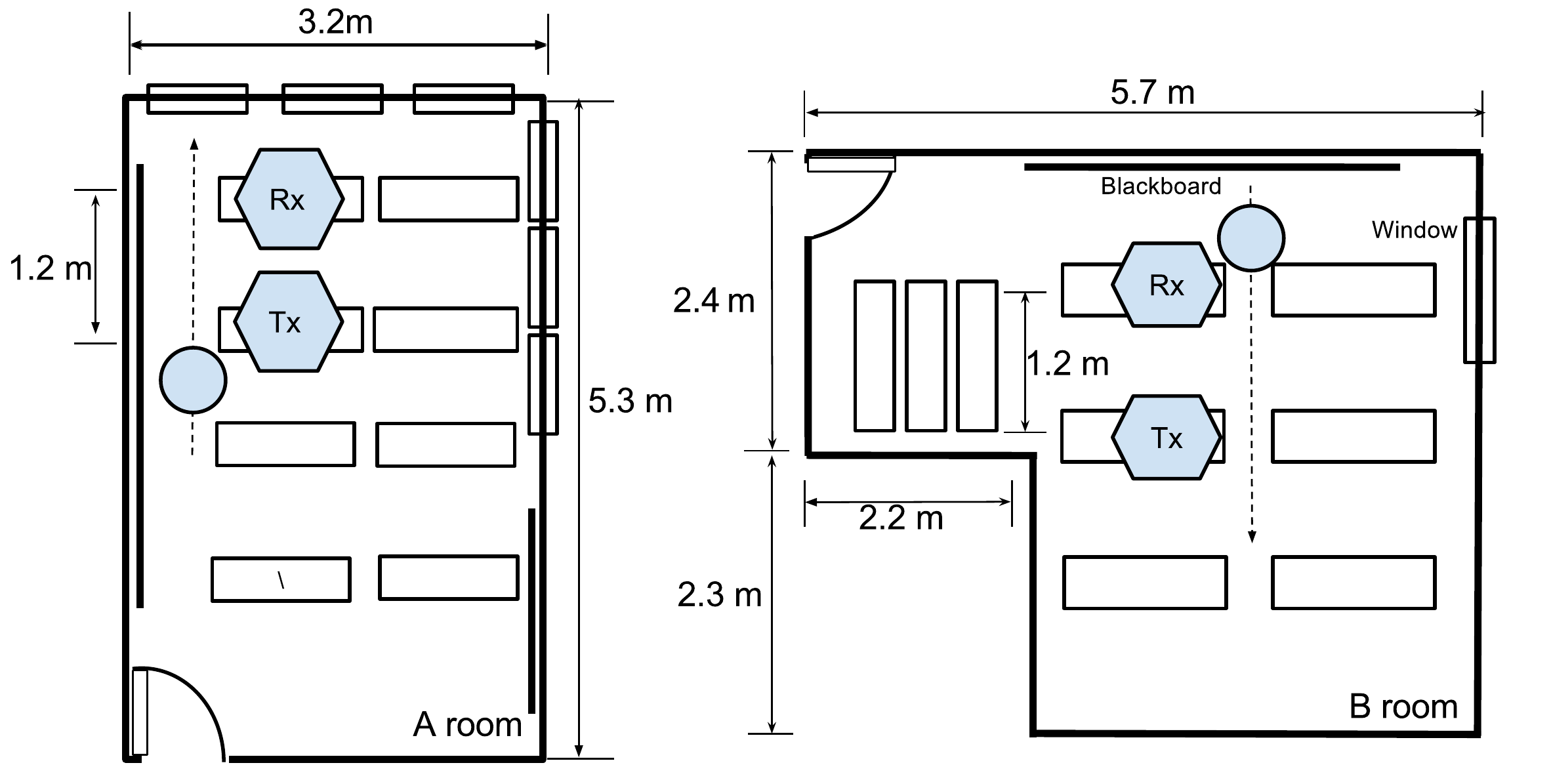}
\caption{Examples of the seminar rooms. Subject performs actions along the trajectories and in circles.}
\label{fig:R539_R540_R542}
}
\end{figure}

 We collect our dataset in five seminar rooms, such as the two shown in
Figure \ref{fig:R539_R540_R542}, to verify if our method could
recognize actions from different places. We define six actions:
\textit{Stand still}, \textit{Walk}, \textit{Run}, \textit{Pick up
box}, \textit{Golf swing}, and \textit{Jump}. A single subject
performs each action 20 times so in total we have 120 traces for each
room. All actions are performed in a 5-second period, 
each generating four sets of CSI with dimension $t \times 30$ and roughly more than 2000 samples/sec. 
We then transform them into four images of size $576 \times 432$.

\begin{table}[t]
\begin{center}
\resizebox{\columnwidth}{!}{
\begin{tabular}{|c|c|c|c|c|c|}
\hline
  Room & A & B & C & D & E \\\hline
  \# of data & 120 & 120 & 120 & 120 & 120 \\\hline
  None-4SVM & 30.00\% & 29.17\% & 26.67\% & 30.00\% & 30.83\% \\\hline
  SVD30-1SVM & 99.17\% & 97.50\% & 97.50\% & 96.67\% & 96.67\% \\
  SVD30-4SVM & 98.33\% & 96.67\% & 99.17\% & 97.50\% & 95.83\% \\
  SVD120-1SVM & 99.17\% & 98.33\% & 99.17\% & 99.17\% & 99.17\% \\
  SVD120-4SVM & 99.17\% & 99.17\% & 98.33\% & 99.17\% & 100.00\% \\\hline
\end{tabular}
}
\caption{Cross validation accuracy of each room. (Both training and testing data are at the same room.)}
\label{tab:one_room}
\end{center}
\end{table}

\subsubsection{Results}
\par For each room, we evaluate the performance by 10-fold cross
validation, as accuracy shown in Table \ref{tab:one_room}. We
experiment on four settings, in which SVD$x$-$y$SVM means applying SVD on
$x$ sub-carriers and training $y$ SVM classifiers. If we do not apply SVD on CSI, performance debases significantly due to noises and background interference, showing that our method can also remove noises. With our setting and de-noise strategy, the performance of
10-fold cross validation in one room is remarkably better than those
without location dependency removal method.

\par Then, we also conduct cross room experiments with one room being
testing data and others being training data, as shown in Table
\ref{tab:cross_room}. The results of none pre-processing set are not
even better than random guess, while those applying SVD achieve a
promising accuracy. Moreover, we can observe that SVD120 outperforms
SVD30 in average performance, inferring that 120 sub-carriers
considered together capture more human action information. However,
the results show that both fusion schemes produce similar performance.

\begin{table}[t]
\begin{center}
\resizebox{\columnwidth}{!}{
\begin{tabular}{|c|c|c|c|c|c|}
\hline
  Testing room & A & B  & C & D & E \\\hline
   None-4SVM & 16.67\% & 16.67\% & 16.67\% & 16.67\% & 16.67\% \\\hline
   SVD30-1SVM & 80.83\% & 89.17\% & 89.17\% & 99.17\% & 98.33\% \\
   SVD30-4SVM & 84.17\% & 53.33\% & 83.33\% & 90.83\% & 96.67\% \\
   SVD120-1SVM & 76.67\% & 91.67\% & 97.50\% & 98.33\% & 95.83\% \\
   SVD120-4SVM & 84.17\% & 75.83\% & 81.67\% & 84.17\% & 98.33\% \\\hline
\end{tabular}
}
\caption{Accuracy of leave-one-out testing (all other rooms as training data).}
\label{tab:cross_room}
\end{center}
\end{table}

\par Finally, we would like to explore whether the amount of training data
affects the accuracy. Since the SVD120-1SVM obtains the better average
performance in the previous experiment, we only list the results using
SVD120-1SVM in Table \ref{tab:more_data}. The results among all rooms
show tendency that when more training data are involved, the higher
accuracy may be achieved, implying the model is improved by
investigating more data. 

\begin{table}[t]
\begin{center}
\resizebox{\columnwidth}{!}{
\begin{tabular}{|c|c|c|c|c|c|}
\hline
  Testing room & A & B & C & D & E \\\hline
  Train with 1 room & 66.67\% & 40.00\% & 43.33\% & 31.67\% & 71.67\% \\
  Train with 2 rooms & 66.67\% & 88.33\% & 95.83\% & 91.67\% & 96.67\% \\
  Train with 3 rooms & 74.17\% & 95.00\% & 95.83\% & 95.83\% & 98.33\% \\
  Train with 4 rooms & 76.67\% & 91.67\% & 97.50\% & 98.33\% & 96.67\% \\\hline
\end{tabular}
}
\caption{Accuracy on SVD120-1SVM of increasing amount of training data (other rooms as training data).}
\label{tab:more_data}
\end{center}
\end{table}

\subsection{Person Identification}
\subsubsection{Dataset and Settings}
\par We invite seven male and three female subjects with age ranging from 20 to 27 and ask each of them to perform walking. All actions are also performed in a 5-second period, and the remaining settings are the same as those in action recognition.

\subsubsection{Results}
\par We separate our experiments into three parts, with results shown in the three blocks of Table \ref{tab:id}. For each part, we follow the same procedure as in action recognition, where we perform 10-fold cross validation on a sub dataset. For the first part, we test on the 100 instances with result demonstrated in the first row, indicating that different person indeed has his or her own walking style, and thus enabling precise identification.

\par In the next part, we investigate if the proposed method could correctly identify the subject given different outfits or accessories. For this purpose, we ask one of the volunteer to record 30 additional data wearing long-sleeve T-shirt, cotton suit and coat. Also, we ask the subject to record 20 additional data bearing side bag or backpack in the original outfit. Hence, in total, we have 50 additional data for the chosen volunteer. We experiment adding data instances of different clothes, different bags or both to our dataset and the minute degradation implies that the classifier could still recognize the person even in different clothes and accessories.

\par In the last part, we deliberately label wrongly the additional ``clothes'' and ``bags'' data of the chosen subject as of 5 different people. That is, we regard each setting as if recorded by a new person. As shown in the last block of the table, the plummeting accuracies verify that if we view instances of the same person with different outfits or accessories as of different people, the classifier would be compromised trying to separate instances that are literally of the same person, leading to a huge debasement in performance.

\begin{table}[t]
\begin{center}
\resizebox{\columnwidth}{!}{
\begin{tabular}{|c|c|c|c|}
\hline
   Used data & \# of data & Accuracy & Remarks \\\hline
   O & 100 & 91\% & Total 10 people \\
   (O + C)$^{\star}$ & 100 & 87\% & \\
   (O + B)$^{\star}$ & 100 & 86\% & \\
   O + C & 130 & 86.43\% & \\
   O + B & 120 & 86.25\% & \\
   O + B + C & 150 & 88.13\% & \\\hline
   C as avatars & 130 & \bf{13.85\%} & Total 13 people \\
   B as avatars & 120 & \bf{18.33\%} & Total 12 people\\
   C and B as avatars & 150 & \bf{14\%} & Total 15 people \\\hline

\end{tabular}
}
\scriptsize{$^{\star}$We randomly sample 10 walks of the chosen subject to match the \# of instances of other subjects\\}
\caption{Cross validation accuracy of identification with different settings using 10 walks per person. In the ``Used data'' column, {\em O} means the original 10-people dataset; {\em C} means 30 additional data of the volunteer wearing three different clothes; and {\em B} means 20 additional data of the volunteer bearing side bag or backpack. We can see that despite different clothing or accessories, our methods could still correctly identify the subjects.}

\label{tab:id}
\end{center}
\end{table}

\section{Discussions and Open Problems}

\par
\begin{figure}[t]
\centering{
\resizebox{\columnwidth}{!}{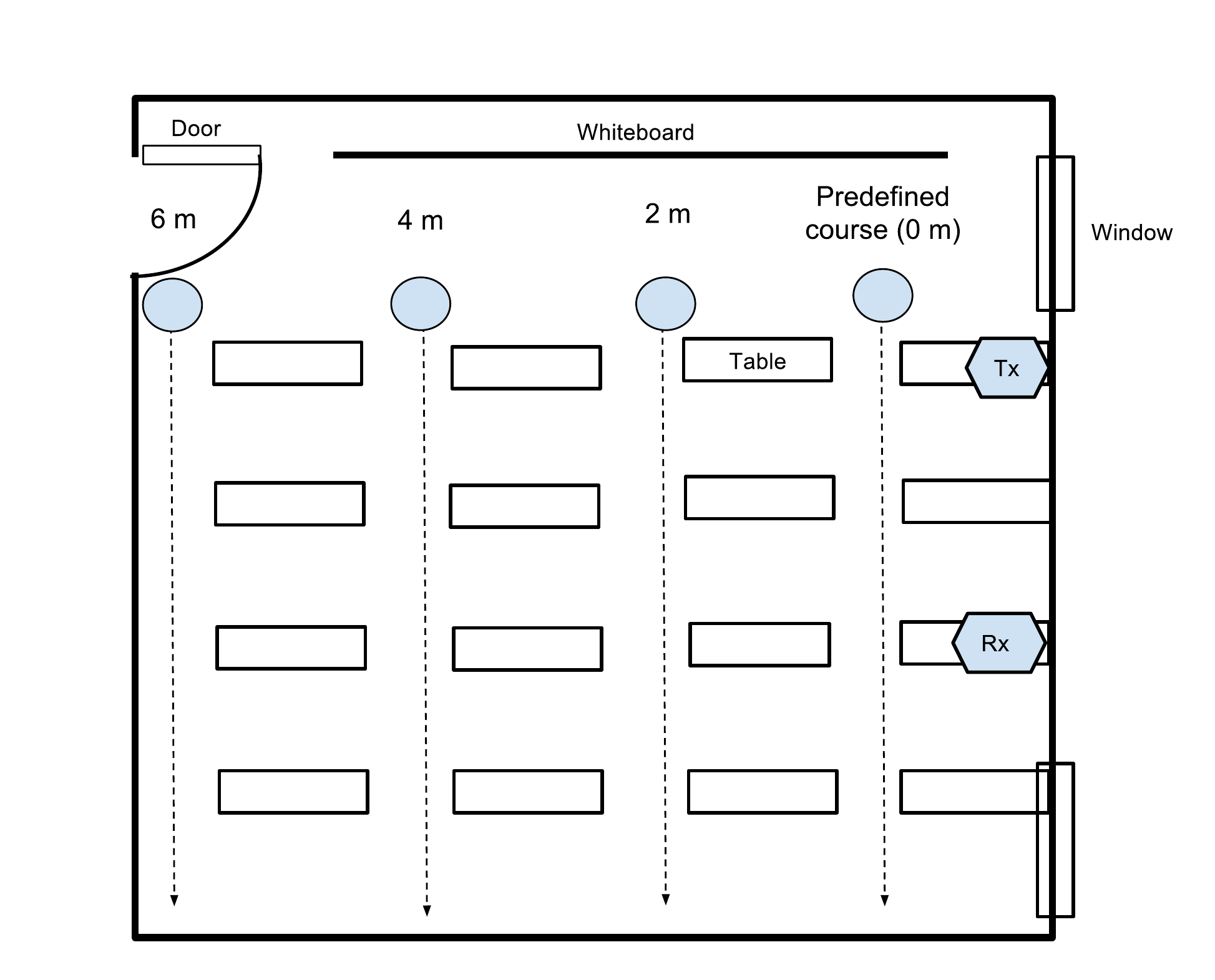}
\caption{Positions and courses for distances mixed experiment.}
\label{fig:course_demo}
}
\end{figure}

 In this section, we study how distances, sampling rates or outfits would affect the performance and find out that only distance has noticeable effect while other factors have minor impact, with results demonstrated in Table \ref{tab:cross_factors}. For each setting (room, distance, sampling rate and distance), we record 60 instances, that is, 10 instances for each action. For the {\em Original} row, we record our dataset at the predefined position/course (as shown in Figure \ref{fig:course_demo}) with sampling rate 2400 samples/sec and T-shirt as our subject$'$s outfit. For the {\em Environments} row, we record an additional dataset for each of the other 4 rooms. Hence, in total, we have 5 datasets of size 60. We then evenly and randomly sample 60 instances from these 5 sets to perform cross validation. For the {\em Distances} row, three additional datasets are recorded 2m, 4m and 6m away from the predefined position/course (please refer to Figure \ref{fig:course_demo} for a clearer picture.) and, same as in {\em Environments}, are randomly sampled 60 instances for cross validation.  For the {\em Sampling rates} row, we again record 3 other datasets with sampling rate 400, 800 and 1200 samples/sec and follow the same procedure as in previous two rows. Lastly for the {\em Outfits} row, 3 datasets are employed with our subject wearing long-sleeve T-shirt, cotton suit and coat.

\par The accuracy is evaluated by 10-fold cross validation with the proposed location dependency removal method. The results show that different distances would debate the accuracy slightly while other factors would not. We consider that even our method removes the environment information, data recorded in various distances would have different human-body-reflected signal power. Thus, proper pre-processing to normalize the power distribution may be worth investigated in the future if we want to recognize data from different distances.

\begin{table}[t]
\begin{center}
\resizebox{\columnwidth}{!}{
\begin{tabular}{|c|c|c|c|c|c|c|c|}
\hline
   - & Stand still & Jump & Pick up box & Run & Golf swing & Walk & All \\\hline
   Original & 100\% & 90\% & 100\% & 95\% & 100\% & 100\% & 97.5 \\\hline
   Environments & 100\% & 85\% & 100\% & 95\% & 85\% & 100\% & 94.17\% \\
   Distances & 75\% & 90\% & 80\% & 95\% & 95\% & 100\% & \bf{89.17\%} \\
   Sampling rates & 95\% & 85\% & 95\% & 95\% & 95\% & 100\% & 94.17\% \\
   Outfits & 100\% & 85\% & 100\% & 100\% & 95\% & 100\% & 96.67\% \\\hline
\end{tabular}
}
\caption{Cross validation accuracy of mixed datasets.}
\label{tab:cross_factors}
\end{center}
\end{table}

\par As for identification, however, we do not conduct experiments of different environments, distances and sampling rates as we did on action recognition. Besides, some open issues could be further discussed, such as detecting a person outside the group, recognizing even a larger group of people, evaluating if diverse age would affect, and on-line learning the classifier adapting to the change in the outfit and moving patterns of a person.

\section{Conclusion}
\par In this paper, we provide an overview on human action recognition along with user identification using Wi-Fi, and clearly differentiate our work with other related researches.

\par We observe the resemblance of CSI to texture and apply vision-based methods on images transformed from CSI. With this brand new angle, we develop a framework that achieves an accuracy above 85\% identifying the seven predefined actions. Though environment dependency is a challenging issue, which lowers the performance when the user deviates from the training locations too much, we propose a location dependency removal method based on SVD to remove the environment information embedded in CSI. Thus, actions recorded in different locations could still be recognized by the trained classifiers. Our experimental results show that the SVD-based solution achieves an accuracy above 90\% classifying the six predefined actions on cross-environment action recognition.

\par Finally, we discuss about the feasibility of our methods applied on identification. Preliminary studies show promising performance identifying people's gait. Some potential issues are listed for further investigation as open research problems.

\ifCLASSOPTIONcaptionsoff
  \newpage
\fi


\begin{thebibliography}{1}

\bibitem{icassp}
J.-Y. Chang, et al. "WiFi action recognition via vision-based methods," 2016 IEEE International Conference on Acoustics, Speech and Signal Processing (ICASSP), 2016, pp. 2782-2786.

\bibitem{mm}
J.-Y. Chang, et al. "Location-Independent WiFi Action Recognition via Vision-based Methods," Proceedings of the 24rd ACM international conference on Multimedia, 2016.


\bibitem{nn-traj}
Y. Shi, et al. "Learning Deep Trajectory Descriptor for action recognition in videos using deep neural networks," 2015 IEEE International Conference on Multimedia and Expo (ICME), 2015, pp. 1-6.

\bibitem{racket-sport}
G. Zhu, et al. "Human Behavior Analysis for Highlight Ranking in Broadcast Racket Sports Video," in IEEE Transactions on Multimedia, vol. 9, no. 6, pp. 1167-1182, Oct. 2007.

\bibitem{codebook}
L. Ballan, et al. "Effective Codebooks for Human Action Representation and Classification in Unconstrained Videos," in IEEE Transactions on Multimedia, vol. 14, no. 4, pp. 1234-1245, Aug. 2012.

\bibitem{accelerometer}
P. Casale, O. Pujol, and P. Radeva. "Human activity recognition from accelerometer data using a wearable device," Iberian Conference on Pattern Recognition and Image Analysis (IbPRIA), 2011, pp. 289-296.

\bibitem{deepface}
Y. Taigman, et al. "Deepface: Closing the gap to human-level performance in face verification," Proceedings of the IEEE Conference on Computer Vision and Pattern Recognition (CVPR), 2014, pp. 1701-1708.

\bibitem{sensorgait}
S.-J. Bamberg, et al. "Gait analysis using a shoe-integrated wireless sensor system," IEEE transactions on information technology in biomedicine, vol. 12, no. 4, 2008, pp. 413-423.

\bibitem{wisee}
Q. Pu, et al. "Whole-home gesture recognition using wireless signals." Proceedings of the 19th annual international conference on Mobile computing \& networking (MobiCom), ACM, 2013, pp. 27-38.

\bibitem{allsee}
R. Nandakumar, B. Kellogg, and S. Gollakota. "Wi-Fi Gesture Recognition on Existing Devices," arXiv preprint arXiv:1411.5394, 2014.

\bibitem{usrp}
Ettus, Matt. "USRP user’s and developer’s guide." Ettus Research LLC (2005).

\bibitem{warp}
P. Murphy, A. Sabharwal, and B. Aazhang. "Design of WARP: a wireless open-access research platform." Signal Processing Conference, IEEE, 2006, pp.1-5.

\bibitem{csitool}
D. Halperin, et al. "Tool release: gathering 802.11 n traces with channel state information," ACM SIGCOMM Computer Communication Review, vol. 41, no. 1, 2011, pp. 53-53.

\bibitem{shopper}
Y. Zeng, P.-H. Pathak, and P. Mohapatra. "Analyzing Shopper's Behavior through WiFi Signals," Proceedings of the 2nd workshop on Workshop on Physical Analytics, ACM, 2015, pp. 13-18

\bibitem{eeye}
Y. Wang, et al.  "E-eyes: device-free location-oriented activity identification using fine-grained WiFi signatures," Proceedings of the 20th annual international conference on Mobile computing and networking (MobiCom), ACM, 2014, pp. 617-628.

\bibitem{wihear}
G. Wang, et al. "We can hear you with wi-fi!," Proceedings of the 20th annual international conference on Mobile computing and networking (MobiCom), ACM, 2014, pp. 593-604.

\bibitem{carm}
W. Wang, et al. "Understanding and modeling of wifi signal based human activity recognition," Proceedings of the 21st Annual International Conference on Mobile Computing and Networking (MobiCom), ACM, 2015, pp. 65-76.

\bibitem{wifall}
C. Han, et al. "WiFall: Device-free fall detection by wireless networks," IEEE INFOCOM 2014-IEEE Conference on Computer Communications, 2014, pp. 271-279

\bibitem{rfcapture}
F. Adib, et al. "Capturing the human figure through a wall," ACM Transactions on Graphics, vol. 34, no. 6, 2015, pp. 219.


\bibitem{freesense}
T. Xin, et al. "FreeSense: Indoor Human Identification with WiFi Signals," IEEE Global Communications Conference (GLOBECOM), 2016, pp. 1-7.

\bibitem{WifiU}
W. Wang, A. Liu, M. Shahzad. "Gait recognition using wifi signals," Proceedings of the 2016 ACM International Joint Conference on Pervasive and Ubiquitous Computing (UbiComp), pp. 363-373, Sep. 2016.

\bibitem{witrack}
F. Adib, et al. "3D tracking via body radio reflections," 11th USENIX Symposium on Networked Systems Design and Implementation (NSDI), 2014, pp. 317-329.


\bibitem{ofdm}
T. Hwang, et al. "OFDM and its wireless applications: a survey," IEEE transactions on Vehicular Technology, vol. 58, no. 4, 2009, pp. 1673-1694.

\bibitem{gabor}
J.-R. Movellan, "Tutorial on Gabor filters," Open Source Document (2002).

\bibitem{sift}
D.-G. Lowe, "Distinctive image features from scale-invariant keypoints," International journal of computer vision, vol. 60, no. 2, 2004, pp. 91-110.

\bibitem{svd}
R.-A. Sadek, "SVD based image processing applications: state of the art, contributions and research challenges," arXiv preprint arXiv:1211.7102, 2012.

\bibitem{KTH}
C. Schuldt, I. Laptev, and B. Caputo. "Recognizing human actions: a local SVM approach," Proceedings of the 17th International Conference on Pattern Recognition (ICPR), IEEE, 2004, pp. 32-36.

\end{thebibliography}
\end{document}